\documentclass[pmlr]{jmlr}

\usepackage{enumitem}
\usepackage{bbold}
\usepackage{listings}

\usepackage{longtable}

\usepackage{booktabs}
\usepackage[load-configurations=version-1]{siunitx} 

\theorembodyfont{\upshape}
\theoremheaderfont{\scshape}
\theorempostheader{:}
\theoremsep{\newline}

\jmlrvolume{230}
\jmlryear{2024}
\jmlrworkshop{Conformal and Probabilistic Prediction with Applications}

\title[Conformal Predictive Systems Under Covariate Shift]{Conformal Predictive Systems Under Covariate Shift}

  \author{\Name{Jef Jonkers\nametag{\thanks{Corresponding author}}} \Email{jef.jonkers@ugent.be} \\
  \addr IDLab, Department of Electronics and Information Systems, Ghent University, Belgium
  \AND
  \Name{Glenn {Van Wallendael}} \Email{glenn.vanwallendael@ugent.be} \\
  \addr IDLab, Department of Electronics and Information Systems, Ghent University - imec, Belgium
  \AND
  \Name{Luc Duchateau} \Email{luc.duchateau@ugent.be} \\
  \addr Biometrics Research Group, Department of Morphology, Imaging, Orthopedics, Rehabilitation and Nutrition, Ghent University, Belgium
  \AND
   \Name{Sofie {Van Hoecke}} \Email{sofie.vanhoecke@ugent.be}\\
   \addr IDLab, Department of Electronics and Information Systems, Ghent University - imec, Belgium}

\editor{Simone Vantini, Matteo Fontana, Aldo Solari, Henrik Boström and Lars Carlsson}

\begin{document}

\maketitle

\begin{abstract}
Conformal predictive systems (CPS) offer a versatile framework for constructing predictive distributions, allowing for calibrated inference and informative decision-making. However, their applicability has been limited to scenarios adhering to the independent and identically distributed (IID) model assumption. This paper extends CPS to accommodate scenarios characterized by covariate shifts. We therefore propose weighted CPS (WCPS), akin to weighted conformal prediction (WCP), leveraging likelihood ratios between training and testing covariate distributions. This extension enables the construction of nonparametric predictive distributions capable of handling covariate shifts. We present theoretical underpinnings and conjectures regarding the validity and efficacy of WCPS and demonstrate its utility through empirical evaluations on both synthetic and real-world datasets. Our simulation experiments indicate that WCPS are probabilistically calibrated under covariate shift.
\end{abstract}

\section{Introduction}
Conformal predictive systems (CPS) are a relatively recent development in Conformal prediction (CP) \citep{vovk_nonparametric_2019, vovk_computationally_2020}. CPS construct predictive distributions by arranging p-values into a nonparametric probability distribution. This distribution satisfies a finite-sample property of validity under the independent and identically distributed (IID) model, i.e., the observations are produced independently from the same probability measure. CPS can be seen as a generalization of point and conformal regressors since they can easily produce point predictions and prediction intervals by leveraging the generated predictive distributions. They allow for more informative and trustworthy decision-making \citep{vovk_conformal_2018}. For example, an obvious use case for predictive distributions is to obtain threshold values such that a decision-maker can determine the likelihood of a certain outcome falling within a specific range. This is particularly useful in applications where understanding the probability distribution of predictions is crucial, such as in risk management, medical diagnostics, and financial forecasting. By setting appropriate thresholds based on the predictive distributions, one can make more informed decisions that balance the trade-offs between different types of errors or risks.

In alignment with the inception of conformal regressors, several adaptations, and enhancements have emerged in the literature after the initial work of \citet{vovk_nonparametric_2019}. These include more computationally efficient variants ~\citep{vovk_computationally_2020}, adaptive versions \citep{vovk_conformal_2020, bostrom_mondrian_2021, johansson_conformal_2023, jonkers_novel_2024}, and proving the existence of universal consistent CPS ~\citep{vovk_universal_2022}.

The exchangeability assumption, which allows for provably valid inference for CP and is a weaker assumption than the IID assumption \citep{shafer_tutorial_2008}, and similarly, the IID assumption for CPS, are standard in machine learning. However, distributional shifts between training and inference data are common in time series, counterfactual inference, and machine learning for scientific discovery but violate these assumptions. While a growing amount of literature has been contributed to extending CP beyond the exchangeability assumptions \citep{tibshirani_conformal_2019, gibbs_adaptive_2021, prinster_jaws_2022, yang_doubly_2022, gibbs_conformal_2023}, allowing (conservatively) valid inference under various types of distributional shifts, no contribution has been made towards extending CPS beyond the IID model. For example, in treatment effect estimation, this extension allows for calibrated predictive distribution beyond the randomized trial setting~\citep{jonkers_conformal_2024}, as in a nonrandomized setting, the covariate distributions for treated and control subjects differ from the target population. Therefore, this work extends CPS beyond the IID model by proposing weighted CPS (WCPS) that constructs valid nonparametric predictive distributions for problems where the covariate distributions of the training and testing data differ, assuming their likelihood ratio is known or can be estimated.

The remainder of this paper is organized as follows: in Section \ref{sec:background}, we will give some background and restate propositions around CP, CPS, and covariate shifts. Section \ref{sec:wcps} presents our modification of CPS to deal with covariate shift, followed by Section \ref{sec:experiments} and Section \ref{sec:conclusion}, which discusses and summarizes the main findings, respectively.

\section{Background}
\label{sec:background}
Let $\textbf{Z} := \textbf{X} \times \mathbb{R}$ be the observation space where each observation $z=(x,y) \in \textbf{Z}$ consist of an object $x \in \textbf{X}$ and its label $y \in \mathbb{R}$. Additionaly, lets $z_1, ..., z_n$ be the training sequence and $z_{n+1} = (x_{n+1}, y_{n+1})$ be the test observation.

\subsection{Conformal Prediction}
Conformal prediction (CP) \citep{vovk_algorithmic_2022} is a model-agnostic and distribution-free framework that allows us to give an implicit confidence estimate in a prediction by generating prediction sets at a specific significance level $\alpha$. The framework provides (conservatively) valid non-asymptotic confidence predictors under the exchangeability assumption. This exchangeability assumption assumes that the training/calibration data should be exchangeable with the test data. The prediction sets in CP are formed by comparing nonconformity scores of examples that quantify how unusual a predicted label is, i.e., these scores measure the disagreement between the prediction and the actual target.

To do so, we define a prediction interval $\hat{C}(x_{n+1})$, for test object $x_{n+1} \in \mathbf{X}$, by calculating following conformity scores $R_i^y$, based on conformity measure $A$, for each $y \in \mathbb{R}$:
\begin{equation}
        R_i^y = A(z_{1:n \setminus i} \cup  (x_{n+1}, y), z_i), \quad i=1,...,n
\end{equation}
and
\begin{equation}
        R_{n+1}^y = A(z_{1:n}, (x_{n+1}, y)).
\end{equation}
The label $y$ is then included in prediction interval $\hat{C}(x_{n+1})$ if,
\begin{equation}
    \frac{|i=1,...,n+1: R_i^y \geq R_{n+1}^y|}{n+1} > \alpha
\end{equation}

The procedure above is referred to as full or transductive conformal prediction and is computationally heavy. Therefore, \citet{papadopoulos_inductive_2002} proposed a more applicable variant of full CP, called inductive or split CP (ICP). ICP is computationally less demanding and allows the use of CP in conjunction with machine learning algorithms, such as neural networks and tree-based algorithms. ICP starts by splitting the training sequence $(x,y) = \{(x_1, y_1), ...,(x_{n}, y_{n})\}$ into a proper training sequence $\{(x_1, y_1), ..., (x_{m}, y_{m})\}$ and a calibration sequence $\{(x_{m+1}, y_{m+1}), ..., (x_{n}, y_{n})\}$. The proper training sequence is used to train a regression model. We then generate nonconformity scores $R_i$ for $(x_i,y_i)$ with $i=m+1,...,n$ from the calibration set, such as for the absolute error, $R_i = |y_i - \hat{y}_i| $. These nonconformity scores are sorted in descending order: $R^*_1,..., R^*_{n-m}$. For a new test object $x_{n+1}$, point prediction $\hat{y}_{n+1}$, and a desired target coverage of $1 - \alpha$, ICP outputs the following prediction interval:
\begin{equation}
    [\hat{y}_{n+1} - R^*_s, \hat{y}_{n+1} + R^*_{s}]
\end{equation}
where $s=\lfloor \alpha (n-m+1) \rfloor$.
\subsection{Covariate Shift}
A covariate shift is a distributional shift where the test object $(x_{n+1}, y_{n+1})$ is differently distributed, i.e. $x_{n+1} \sim \tilde{P}_{X}$, than the training data $z_i = (x_i, y_i), i=1,...,n$ where $x_i \sim P_{X}$, thus $\tilde{P}_{X} \neq P_{X}$. However, the relationship between inputs and labels remains fixed.
\begin{equation}
\label{eq:cov-shift}
\begin{split}
    (x_i, y_i) \mathop{\sim}\limits^{\mathrm{iid}} P &= P_{X} \times P_{Y|X}, \quad i=1,...,n \\
    (x_{n+1}, y_{n+1}) \sim \tilde{P} &= \tilde{P}_{X} \times P_{Y|X}
\end{split}
\end{equation}
\subsection{Weighted Conformal Prediction}
\citet{tibshirani_conformal_2019} was one of the first works to extend conformal prediction beyond the exchangeability assumption to deal with covariate shifts. Specifically, they propose weighted conformal prediction (WCP) to deal with covariate shifts where the likelihood ratio between the training $P_X$ and test $\tilde{P}_X$ covariate distributions is known. In WCP, the empirical distribution of nonconformity scores at the calibration points gets reweighted, and thus each nonconformity score $R_i$ gets weighted by a probability $p_i^w(x)$ proportional to the likelihood ratio $w(x_i) = \frac{d\tilde{P}_X(x_i)}{dP_X(x_i)}$:
\begin{align}
    p_i^w(x) &= \frac{w(x_i)}{\sum_{j=1}^n w(x_j) + w(x)}, \qquad i=1,...,n, \\
    p_{n+1}^w(x) &= \frac{w(x)}{\sum_{j=1}^n w(x_j) + w(x)}.
\end{align}
This results in an adjusted empirical distribution of nonconformity scores depicted in Table \ref{tab:emp-dist}.
\begin{table}[hbtp]
\floatconts
  {tab:emp-dist}
  {\caption{Empirical distribution of nonconformity scores ($\delta_a$ denotes a point mass at $a$).}}
  {\begin{tabular}{c|c}
  \toprule
  \bfseries Regular & \bfseries Weighted\\
  \midrule
  $\frac{1}{n+1} \sum_{i=1}^n \delta_{R_i} + \frac{1}{n+1}\delta_{\infty}$ & $\sum_{i=1}^n p_i^w(x) \delta_{R_i} + p_{n+1}^w(x) \delta_{\infty}$ \\
  \bottomrule
  \end{tabular}}
\end{table}
\citet{tibshirani_conformal_2019} showed that the validity of WCP remains even for the computational less-demanding split conformal prediction. However, this all does not come for free; we are reducing the sample size by weighting nonconformity scores and consequentially losing some reliability, i.e., variability in empirical coverage, compared to CP without covariate shift and the same number of samples. \citet{tibshirani_conformal_2019} pointed out a popular heuristic from the covariate shift literature \citep{gretton_covariate_2008, reddi_doubly_2015} to determine the effective sample size $\hat{n}$ of $X_1, ..., X_n$ training points, and a likelihood ratio $w$:
\begin{equation}
\label{eq:effective-sample}
    \hat{n} = \frac{[\sum_{i=1}^n |w(x_i)|]^2}{\sum_{i=1}^n |w(x_i)|^2} = \frac{||w(x_{1:n})||_1^2}{||w(x_{1:n})||_2^2}
\end{equation}
where $w(x_{1:n}) = (w(x_1), ..., w(x_n))$. 
Note that it is possible to learn the likelihood ratio $w(x_i) = \frac{d\tilde{P}_X(x_i)}{dP_X(x_i)}$ between training and test covariate distribution, as showed by \citet{tibshirani_conformal_2019}, if it is reasonably accurate.

\subsection{Conformal Predictive Systems}
Conformal predictive systems (CPS) allow the construction of predictive distributions by extending upon full CP. CPS produces conformal predictive distributions by arranging p-values into a probability distribution function~\citep{vovk_nonparametric_2019}. These p-values are created with the help of specific types of conformity measures. \citet{vovk_nonparametric_2019} defines a CPS as a function that is both a conformal transducer (Definition \ref{def:ct}) and a randomized predictive system (RPS) (Definition \ref{def:RPS}).
\begin{definition}[Conformal Transducer, \cite{vovk_algorithmic_2022}]
\label{def:ct}
   The conformal transducer determined by a conformity measure $A$ is defined as,
   \begin{equation*}
        Q(z_1,...,z_n,(x_{n+1},y), \tau) := \sum_{i=1}^{n+1} [R_i^y < R_{n+1}^y] \frac{1}{n+1} + \sum_{i=1}^{n+1} [R_i^y = R_{n+1}^y] \frac{\tau}{n+1} 
   \end{equation*}
   where $(z_1, ..., z_n)$ is the training sequence, $\tau \in [0,1]$, $x_{n+1}$ is a test object, and for each label $y$ the corresponding conformity score $R_i^y$ is defined as
   \begin{equation*}
        \begin{split} 
            R_i^y &:= A(z_1, ...,z_{i-1}, z_{i+1}, ..., z_{n}, (x_{n+1}, y), z_i), \qquad i = 1, ..., n \\
            R_{n+1}^y &:= A(z_1, ..., z_{n}, (x_{n+1}, y)).
        \end{split}
    \end{equation*}
\end{definition}

\begin{definition}[RPS, \cite{vovk_nonparametric_2019}]\label{def:RPS}
A function $Q: \textbf{Z}^{n+1} \times [0,1] \rightarrow [0,1]$ is an RPS if it satisfies the following requirements:
\begin{itemize}
    \item [R1.1] For each training sequence $(z_1, ..., z_n) \in \textbf{Z}^n$ and test object $x \in \textbf{X}$, the function $Q(z_1, ..., z_n, (x_{n+1}, y), \tau)$ is monotonically increasing both in $y$ and $\tau$. Put differently, for each $\tau \in [0, 1]$, the function
        \begin{equation*}
            y \in \mathbb{R} \rightarrow Q(z_1, ..., z_n, (x_{n+1}, y), \tau)
        \end{equation*}
    is monotonically increasing, and for each $y \in \mathbb{R}$, the function
        \begin{equation*}
            \tau \in [0, 1] \rightarrow Q(z_1, ..., z_n, (x_{n+1}, y), \tau)
        \end{equation*}
    is also monotonically increasing.
    \item [R1.2] For each $\tau, \tau' \in [0,1]$ and each test object $x_{n+1} \in \textbf{X}$, 
    \begin{equation*}
        Q(z_1, ..., z_n, (x_{n+1}, y), \tau) \geq Q(z_1, ..., z_n, (x_{n+1}, y'), \tau'), \qquad \text{if} \quad y > y'
    \end{equation*}
    \item [R1.3]  For each training sequence $(z_1, ..., z_n) \in \textbf{Z}^n$ and test object $x_{n+1} \in \textbf{X}$,
    \begin{equation*}
        \lim_{y \rightarrow -\infty} Q(z_1, ..., z_n, (x_{n+1}, y), 0) = 0
    \end{equation*}
    and
    \begin{equation*}
        \lim_{y \rightarrow \infty} Q(z_1, ..., z_n, (x_{n+1}, y), 1) = 1
    \end{equation*}
    \item [R2] As a function of random training observations $z_1 \sim P, ..., z_{n} \sim P$, $z_{n+1} \sim P$, and a random number $\tau \sim Uniform(0,1)$, all assumed to be independent, the distribution of Q is uniform:
    \begin{equation*}
        \forall \alpha \in [0,1]: \mathbb{P}\{Q(z_1, ...,z_n, z_{n+1}, \tau) \leq \alpha\} = \alpha 
    \end{equation*}
\end{itemize}
\end{definition}

Definition \ref{def:RPS} that defines an RPS is in verbatim from \cite{vovk_nonparametric_2019}, except requirement \textit{R1.2}, which is appended to the definition as we believe this is a requirement which is implicitly assumed by \citet{vovk_nonparametric_2019}. 

Note that a conformal transducer satisfies \textit{R2} by its validity property (see Proposition 2.11 in \cite{vovk_algorithmic_2022}). Additionally, in \citet{vovk_universal_2022} (Lemma 1), they show that a conformal transducer defined by a monotonic conformity measure $A$ is also an RPS and thus a CPS if $A$ follows the following three conditions:
\begin{itemize}[noitemsep]
    \item for all $n$, all training data sequences $(z_1, ..., z_n)$, and all test objects $x_{n+1}$,
    \begin{equation}
        \label{eq:inf-conj}
        \inf_y A(z_1,...,z_n,(x_{n+1},y)) = \inf A_n
    \end{equation}
    \begin{equation}
        \label{eq:sup-conj}
        \sup_y A(z_1,...,z_n,(x_{n+1},y)) =  \sup A_n;
    \end{equation}
    \item for each n, the $\inf_y$ in Equation \ref{eq:inf-conj} is either attained for all $(z_1, ...,z_n)$ and $x_{n+1}$, or not attained for any $(z_1,...,z_n)$ and $x_{n+1}$;
    \item for each n, the $\sup_y$ in Equation \ref{eq:sup-conj} is either attained for all $(z_1,...,z_n)$ and $x_{n+1}$, or not attained for any $(z_1,...,z_n)$ and $x_{n+1}$.
    \end{itemize}

\subsubsection{Split Conformal Predictive Systems}
Like CP, CPS has been adapted and made more computationally efficient by building upon ICP, namely split conformal predictive systems (SCPS) \citep{vovk_computationally_2020}. Here, the p-values are created by a split conformity measure that needs to be isotonic and balanced. A good and standard choice of split conformity measure, according to \citet{vovk_computationally_2020}, is a (normalized) residual. In Appendix \ref{apd:scps}, we present and discuss, similarly as for CPS, definitions and propositions related to SCPS.

\section{Weighted Conformal Predictive System}
\label{sec:wcps}
As WCP extends upon CP, we propose to reweigh the conformity scores with a probability $p_i^w(x)$ proportional to the likelihood ratio $w(x_i)=\frac{d\tilde{P}_X(x_i)}{dP_X(x_i)}$, to present a weighted conformal transducer where the output is defined by conformity measure $A$ and likelihood ratio $w(x) = \frac{d\tilde{P}_X(x)}{dP_X(x)}$,
\begin{equation}
\label{eq:WCPS}
    Q(z_1,...,z_n,\frac{d\tilde{P}}{dP},(x_{n+1},y), \tau) := \sum_{i=1}^{n+1} [R_i^y < R_{n+1}^y] p_i^w(x) + \sum_{i=1}^{n+1} [R_i^y = R_{n+1}^y] p_i^w(x) \tau 
\end{equation}
where $\tau$ is a random number sampled from a uniform distribution between 0 and 1. Note that under the absence of a covariate shift, the probability weights become equal, $p_i^w(x) = p^w_{n+1} = \frac{1}{n+1}$. In this scenario, the weighted conformal transducer (\ref{eq:WCPS}) will become equivalent to a conformal transducer.

A function is a weighted conformal predictive system (WCPS) if it is both a weighted conformal transducer and a weighted RPS (WRPS), i.e., an RPS probabilistically calibrated under covariate shift. To prove that under specific conformity measures $A$, e.g., monotonic conformity measures, a weighted conformal transducer is also a WRPS, we need to prove Conjecture \ref{conj:valid-wct}, i.e., that the weighted conformal transducer is probabilistically calibrated.
\begin{conjecture}
\label{conj:valid-wct}
Assume that
\begin{itemize}[noitemsep]
    \item $z_i = (x_i, y_i) \in \mathbf{X} \times \mathbb{R}$, $i=1,...,n$ are produced independently from $P = P_X \times P_{Y|X}$;
    \item  $z_{n+1} = (x_{n+1}, y_{n+1}) \in \mathbf{X} \times \mathbb{R}$, is independently drawn from $\tilde{P} = \tilde{P}_X \times P_{Y|X}$;
    \item $\tilde{P}_X$ is absolutely continuous with respect to $P_X$;
    \item random number $\tau \sim Uniform(0,1)$;
    \item $z_{1:n}$, $z_{n+1}$, and $\tau$ to be independent.
\end{itemize}
Then the distribution of the weighted conformal transducer, defined by (\ref{eq:WCPS}), is uniform:
\begin{equation}
    \forall \alpha \in [0,1]: \mathbb{P}_{z_{1:n} \sim P, z_{n+1} \sim \tilde{P}}\{Q(z_1,...,z_n,\frac{d\tilde{P}}{dP},z_{n+1}, \tau) \leq \alpha\} = \alpha
\end{equation}
\end{conjecture}
We leave this proof for future work. However, if proven, Conjecture \ref{conj:WCPS} can be easily proven by following the same procedure as the proof of Lemma 1 in \cite{vovk_universal_2022} using Conjecture \ref{conj:valid-wct} instead of the property of validity of a conformal transducer.

\begin{conjecture}[Weighted Version of Lemma 1 in \cite{vovk_universal_2022}]
\label{conj:WCPS}
    Suppose a monotonic conformity measure $A$ satisfies the following three conditions:
    \begin{itemize}[noitemsep]
        \item for all $n$, all training data sequences $(z_1, ..., z_n)$, and all test objects $x_{n+1}$,
        \begin{equation}
            \label{eq:inf-conj-weighted}
            \inf_y A(z_1,...,z_n,(x_{n+1},y)) = \inf A_n
        \end{equation}
        \begin{equation}
            \label{eq:sup-conj-weighted}
            \sup_y A(z_1,...,z_n,(x_{n+1},y)) =  \sup A_n;
        \end{equation}
        \item for each n, the $\inf_y$ in Equation \ref{eq:inf-conj-weighted} is either attained for all $(z_1, ...,z_n)$ and $x_{n+1}$ or not attained for any $(z_1,...,z_n)$ and $x_{n+1}$;
        \item for each n, the $\sup_y$ in Equation \ref{eq:sup-conj-weighted} is either attained for all $(z_1,...,z_n)$ and $x_{n+1}$ or not attained for any $(z_1,...,z_n)$ and $x_{n+1}.$
    \end{itemize}
    Then, the weighted conformal transducer corresponding to $A$ is a WRPS.
\end{conjecture}
In other words, a weighted conformal transducer based on a monotonic conformity measure satisfying the aforementioned requirements is also a WRPS.


\subsection{Weighted Split Conformal Predictive Systems}
Besides bringing WCPS to CPS, we also propose a more computationally efficient approach to construct calibrated predictive distribution based on SCP by presenting a weighted split conformal transducer determined by the split conformity measure $A$ and likelihood ratio $w(x)$,
\begin{equation}
    \label{eq:split-wct}
    Q(z_1,...,z_n, \frac{d\tilde{P}}{dP}, (x,y), \tau) := \sum_{i=m+1}^{n} [R_i < R^y] p_i^w(x) + \sum_{i=m+1}^{n} [R_i = R^y] p_i^w(x) \tau + p^w_{n+1}(x) \tau
\end{equation}
Similarly to WCPS, a function is a weighted split conformal predictive system (WSCPS) if it is both a weighted split conformal transducer and a WRPS. Thus, we also need to prove a notion of validity in the form of calibration in probability, see Conjecture \ref{conj:valid-split-wct}. We leave this proof for future work, but we show in Section \ref{sec:experiments} with simulation experiments that this empirically seems to be the case.

\begin{conjecture}
\label{conj:valid-split-wct}
Assume that
\begin{itemize}[noitemsep]
    \item the training sequence $z_1,...,z_n$ is split into two parts: the proper training sequence $z_1,...,z_m$ and the calibration sequence $z_{m+1},...,z_n$;
    \item $z_i = (x_i, y_i) \in \mathbb{R}^d \times \mathbb{R}$, $i=m+1,...,n$ are produced independently from $P = P_X \times P_{Y|X}$;
    \item  $z_{n+1} = (x_{n+1}, y_{n+1}) \in \mathbf{X} \times \mathbb{R}$, is independently drawn from $\tilde{P} = \tilde{P}_X \times P_{Y|X}$;
    \item $\tilde{P}_X$ is absolutely continuous with respect to $P_X$;
    \item random number $\tau \sim Uniform(0,1)$;
    \item $z_{m+1:n}$, $z_{n+1}$, and $\tau$ to be independent.
\end{itemize}
Then is the distribution of weighted split conformal transducer, defined by (\ref{eq:split-wct}), uniform:
\begin{equation}
    \forall \alpha \in [0,1]: \mathbb{P}_{z_{m+1:n} \sim P, z_{n+1} \sim \tilde{P}}\{Q(z_1,...,z_n,\frac{d\tilde{P}}{dP},z_{n+1}, \tau) \leq \alpha\} = \alpha
\end{equation}
\end{conjecture}

\begin{conjecture}
    \label{conj:WSCPS}
    The weighted split conformal transducer (\ref{eq:split-wct}) is a WRPS if and only if it is based on a balanced isotonic split conformity measure.
\end{conjecture}

A proof of Conjecture \ref{conj:WSCPS} will follow the same procedure as the proof of Proposition  1 and 2 in \cite{vovk_computationally_2020} using Conjecture \ref{conj:valid-split-wct} instead of the property of validity of a split conformal transducer.

\section{Experiments}
\label{sec:experiments}
We evaluate (weighted) CPS under a covariate shift on empirical and synthetic data, and use (weighted) split CPS approaches for efficiency. For implementing WSCPS, we made an extension of the python package \texttt{crepes} \citep{bostrom_crepes_2022}, named \texttt{crepes-weighted}. A more detailed description can be found in Appendix \ref{apd:package}. The Python code to reproduce the simulation results can be found at \url{https://github.com/predict-idlab/crepes-weighted}.

\subsection{Data}
\subsubsection{Empirical Data}
We consider the airfoil dataset from the UCI Machine Learning Repository \citep{dua_uci_2017}, which contains N = 1503 observation, where each observation consists of a response value Y (scaled sound pressure level of NASA airfoils) and a vector of covariates X with dimension 5 (log frequency, angle of attack, chord length, free-stream velocity, and suction side log displacement thickness). We use the same experimental setting as \citet{tibshirani_conformal_2019} to demonstrate the use of CPS under covariate shifts.

In total, we run 1000 experimental trials. For a single trial, the dataset is split into three sets $D_{train}$, $D_{cal}$, $D_{test}$, which are IID and respectively contain 25\%, 25\%, and 50\% of the data and have the following roles: 
\begin{itemize}[noitemsep]
    \item $D_{train}$ is used as proper training dataset for the CPS, i.e., to train a regression model $\hat{\mu}$.
    \item $D_{cal}$ is used as calibration set to create conformity scores, we will use the residual as conformity measure.
    \item $D_{test}$ is used as our test set and has no covariate shift compared to the other sets.
\end{itemize}

To simulate a covariate shift, \cite{tibshirani_conformal_2019} propose to construct a fourth set $D_{shift}$ that samples with replacement from $D_{test}$, with probabilities proportional to
\begin{equation}
    w(x) = \exp(x^T \beta), \qquad \text{where} \quad \beta = (-1,0,0,0,1).
\end{equation}
We can view $w(x)$ as the likelihood ratio of covariate distributions between the shifted test set $D_{shift}$ and training set $D_{train}$, since $D_{train}$ and $D_{test}$ follow the same IID model. Consequentially, $w(x)$ is used to account for the covariate shift when using a WSCPS.

\subsubsection{Synthetic Data}
We also evaluate our approach on synthetic data to evaluate the assumed validity property, i.e., calibrated in probability, of the WSCPS. We use the setting from \cite{kang_demystifying_2007}, which is also used in \citet{yang_doubly_2022}, where each observation $i$ is generated in the following way:
\begin{itemize}[noitemsep]
    \item $(x_{i1}, x_{i2}, x_{i3}, x_{i4})^T$ is independently distributed as $N(0,I_4)$ where $I_4$ represents the $4 \times 4$ identity matrix.
    \item $y_i = 210 + 27.4 x_{i1} + 13.7 x_{i2} +  13.7 x_{i3} +  13.7 x_{i4} + \varepsilon_i, \qquad \text{where} \quad \varepsilon_i \sim N(0,1)$
    \item $w(x) = \exp(-x_{i1} + 0.5 x_{i2} - 0.25 x_{i3} - 0.1 x_{i4}$), which represents the likelihood ratio of the covariate distributions of the shifted test set $D_{shift}$ and training set $D_{train}$.
\end{itemize}
We also run 1000 experimental trials for the synthetic data experiments.

\subsection{Results}
To evaluate the proposed WSCPS, we perform three different experiments on the empirical and synthetic data. These evaluate the coverage of WSCPS-generated prediction intervals, the quality of predictive distributions, and probabilistic calibration under covariate shift.

First, we evaluate the coverage of 80\% prediction intervals generated with CPS under the IID model and covariate shift, similarly as \cite{tibshirani_conformal_2019} for CP. We can construct prediction intervals by extracting specific percentiles from the conformal predictive distributions, e.g., the 10th and 90th percentile, which are the lower and upper bound of the 80\% prediction interval.

Next, we evaluate the performance of the predictive distributions generated by CPS under the IID model and covariate shift. We consider the continuous ranked probability score (CRPS) to evaluate this, as it is a proper scoring rule for probabilistic forecasting \citep{gneiting_strictly_2007, gneiting_probabilistic_2007}. The CRPS is defined as
\begin{equation}
    CRPS(F, y_i) = \int_{-\infty}^{\infty} (F(y) - \mathbb{1}_{\{y \geq y_i\}})^2 dy
\end{equation}
where $F$ is the distribution function $F: \mathbb{R} \rightarrow [0,1]$, $y_i$ is the observed label, and $\mathbb{1}$ represents the indicator function. The CRPS most minimal value, 0, is achieved when all probability of the predictive distribution is concentrated in $y_i$. Otherwise, the CRPS will be positive. Since SCPS and WSCPS are somewhat fuzzy, the CRPS cannot be computed directly. Therefore, we use the modification of SCPS, proposed by \citet{vovk_computationally_2020}, and adapt it to WSCPS, which ignores the fuzziness represented by the random variable $\tau \sim Uniform(0,1)$.

Finally, we validate by simulation Conjecture \ref{conj:valid-split-wct} by producing p-values with the (W)SCPS by setting $y$ to the label $y_{n+1}$ and checking if their histogram follows a uniform distribution. In the probabilistic forecasting literature, this is often referred to as probability integral transforms (PIT) histograms \citep{gneiting_probabilistic_2007}.

\paragraph{Coverage of intervals under covariate shift} The results are depicted in Figure \ref{fig:cov}. We observe similar results as WCP \citep{tibshirani_conformal_2019}; in row 1 of Figure \ref{fig:cov}) we observe undercoverage for SCPS under covariate shift. The WSCPS brings the average coverage to the desired level under covariate shift for both experiments, while the SCPS constructed intervals considerably undercover; see row 2 of Figure \ref{fig:cov}. We also observe that the heuristic for the reduced (effective) calibration set size due to the weighting operation of WCP, see Equation \ref{eq:effective-sample}, is also a good heuristic for WSCPS. This is shown in the third row of Figure \ref{fig:cov}, where we observe similar dispersion of coverage over experiment trials for WSCPS and SCPS with a reduced calibration set.
\begin{figure}[htbp]
\floatconts
  {fig:cov}
  {\caption{Empirical coverage of 80\% prediction intervals from (W)SCPS, computed using 1000 different random splits of the airfoil and synthetic dataset.}}
  {%
    \subfigure[Airfoil data]{\label{fig:cov-airfoil}%
      \includegraphics[width=0.47\linewidth]{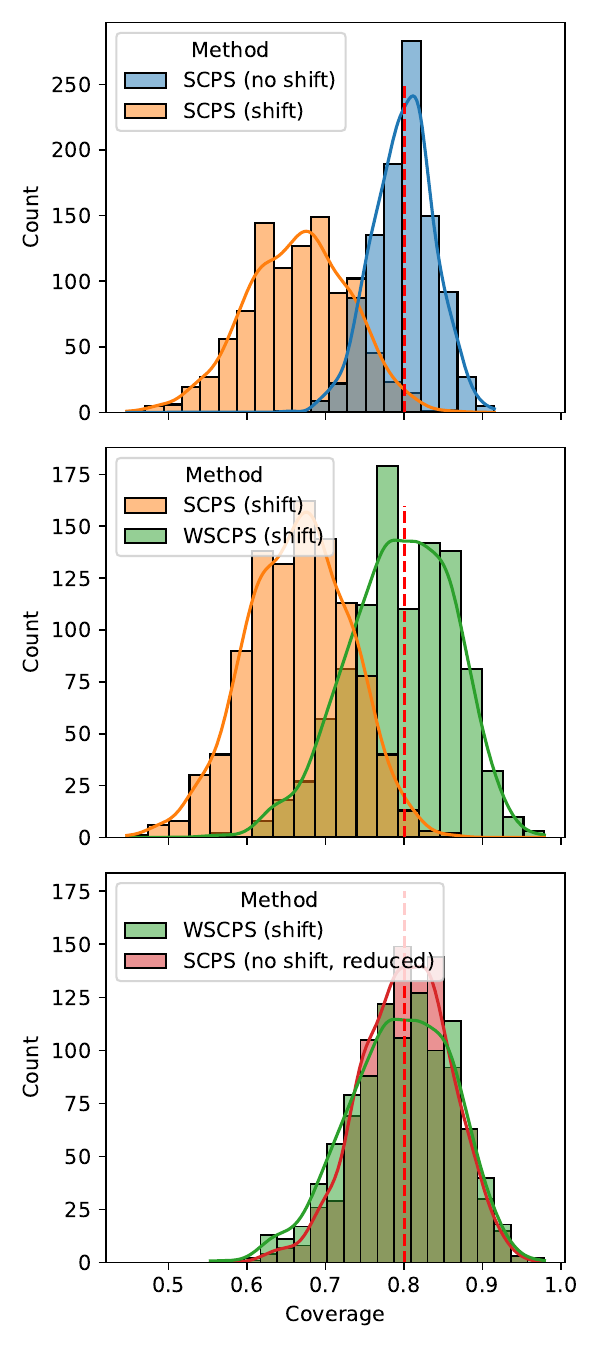}}%
    \qquad
    \subfigure[Synthetic data]{\label{fig:cov-synthetic}%
      \includegraphics[width=0.47\linewidth]{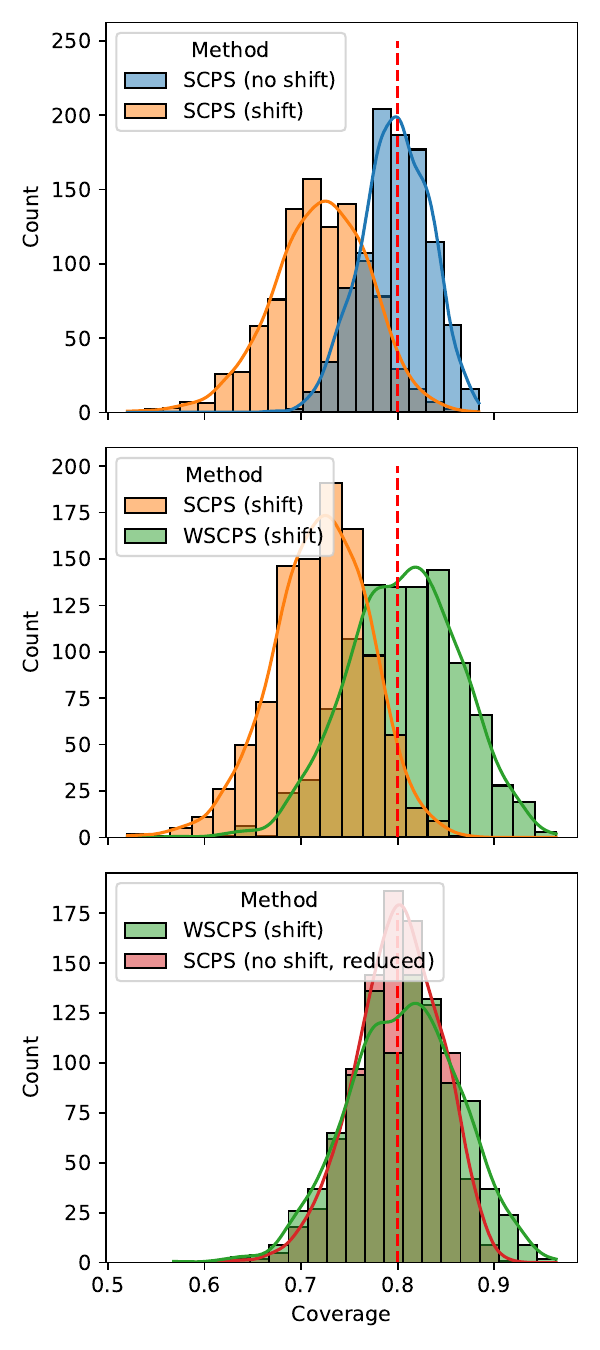}}
  }
\end{figure}
\paragraph{Quality of predictive distribution under covariate shift} Figure \ref{fig:crps-hist} shows the performance of different SCPS in terms of CRPS across the different trials. We see a performance difference when a covariate shift is present and not. The WSCPS consistently (slightly) outperforms the SCPS under covariate shift for both datasets. However, it is difficult to see in the second row of Figure \ref{fig:crps-hist}. Therefore, we also perform a post-hoc Friedman-Nemenyi test (see Figure \ref{fig:crps-tests}). The SCPS under no shift with a calibration set size equal to the effective sample size of WSCPS has a significantly better CRPS score than WSCPS. This is expected since under covariate shift, the model $\hat{\mu}$ is trained on training data differently distributed as the test set, as \citet{tibshirani_conformal_2019} also indicated. Ideally, $\hat{\mu}$ should be adjusted for the covariate shift; however, we leave this for future work. 
\begin{figure}[htbp]
\floatconts
  {fig:crps-hist}
  {\caption{Empirical CRPS of (W)SCPS, computed using 1000 different experiment trials for both airfoil and synthetic datasets.}}
  {%
    \subfigure[Airfoil data]{\label{fig:crps-hist-airfoil}%
      \includegraphics[width=0.47\linewidth]{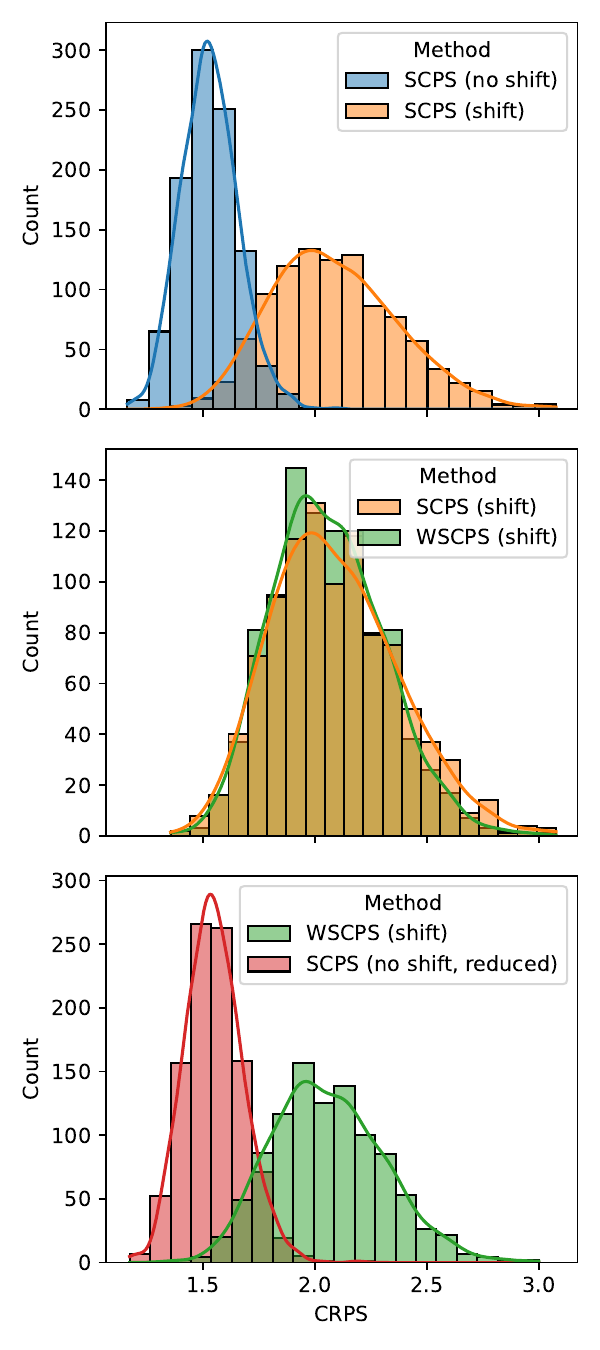}}%
    \qquad
    \subfigure[Synthetic data]{\label{fig:crps-hist-synthetic}%
      \includegraphics[width=0.47\linewidth]{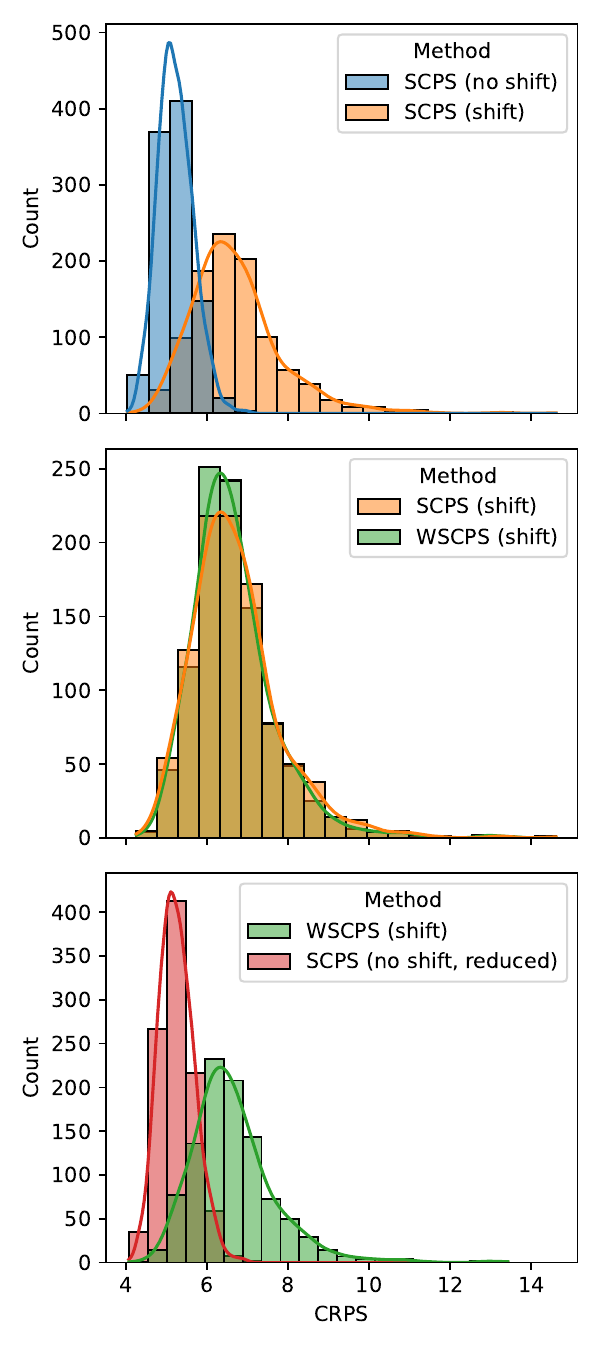}}
  }
\end{figure}

\begin{figure}[htbp]
\floatconts
  {fig:crps-tests}
  {\caption{Post-hoc Friedman-Nemenyi test for CRPS.}}
  {%
    \subfigure[Airfoil data]{\label{fig:crps-tests-airfoil}%
      \includegraphics[width=\linewidth]{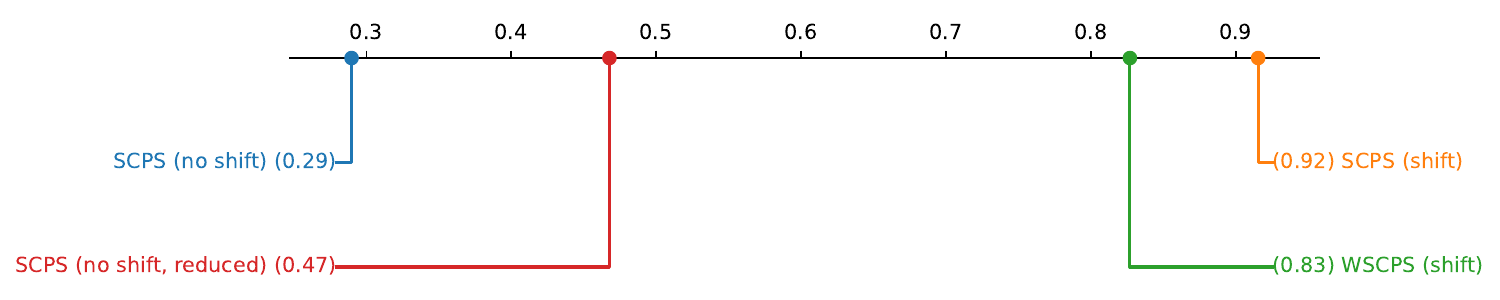}}%
    \qquad
    \subfigure[Synthetic data]{\label{fig:crps-tests-synthetic}%
      \includegraphics[width=\linewidth]{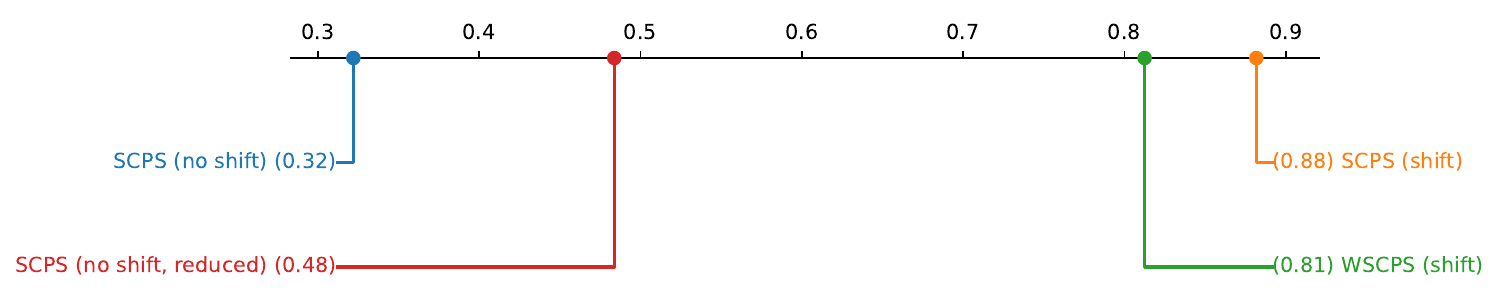}}
  }
\end{figure}

\begin{figure}
    \centering
    \includegraphics[width=\linewidth]{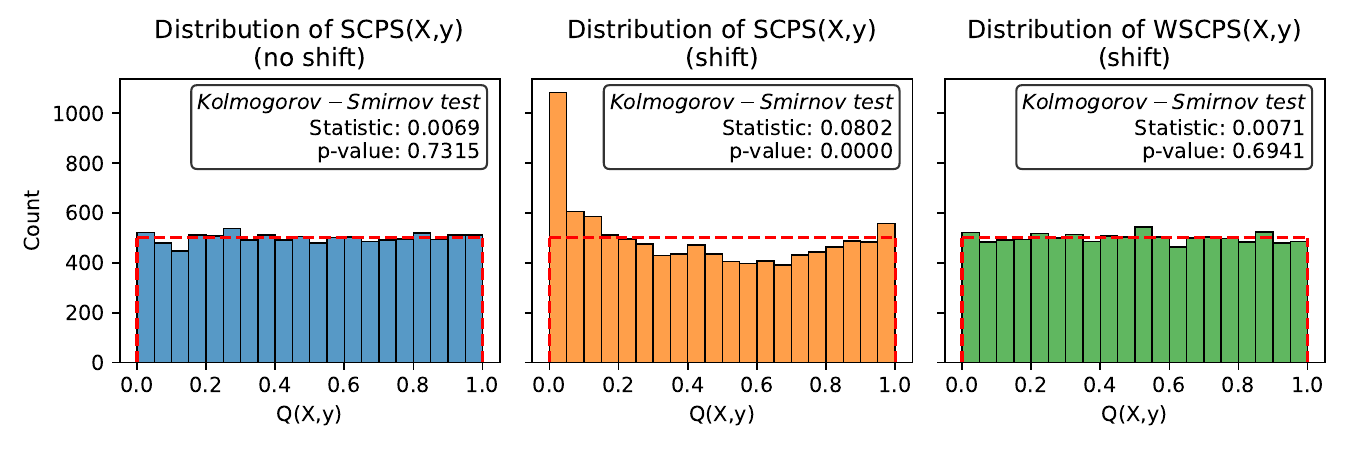}
    \caption{Distribution of p-values of SCPS under IID model (blue), covariate shift (orange), and WSCPS (green). The red dashed line represents the uniform distribution the p-values need to follow so that the (W)SCPS is probabilistically calibrated.}
    \label{fig:dist-p-values}
\end{figure}

\paragraph{Probabilistic calibration under covariate shift} We validate by simulation Conjecture \ref{conj:valid-split-wct}, which states that under covariate shift, the weighted split conformal transducer produced $p$-values are distributed uniformly on $[0,1]$ when we know the likelihood ratio of the covariate distribution of the training and test set. The results of the simulation experiments, depicted in Figure \ref{fig:dist-p-values}, indicate that Conjecture \ref{conj:valid-split-wct} is empirically valid and that it breaks when we do not account for the covariate shift.

\section{Conclusion}
\label{sec:conclusion}
We have introduced a novel extension to conformal predictive systems (CPS) to address covariate shifts in predictive modeling. Covariate shifts are a common challenge in real-world machine learning applications. Our proposed approach, weighted (split) Conformal predictive systems (W(S)CPS), leverages the likelihood ratio between training and testing data distributions to construct calibrated predictive distributions.

We outlined the theoretical framework of WCPS and WSCPS, demonstrating their formal definition and properties. Similarly, as \citet{tibshirani_conformal_2019}, we built upon the foundation of CPS and extended the concept to handle covariate shifts effectively. Our theoretical analysis included conjectures regarding the probabilistic calibration of WCPS under covariate shift, paving the way for future research in this area. Additionally, we successfully validated these conjectures with simulation experiments.

In future work, we aim to provide rigorous proofs for the conjectures presented in this paper to establish the theoretical underpinnings of our proposed methods. Additionally, we will evaluate our proposed framework for counterfactual inference and incorporate it into our recently proposed conformal Monte-Carlo meta-learners \citep{jonkers_conformal_2024}, which opens the possibility of giving validity guarantees for predictive distributions of individual treatment effect beyond the randomized trial setting. Moreover, we believe that similar applications and extensions of the WCP \citep{tibshirani_conformal_2019}, such as addressing label shift \citep{podkopaev_distribution-free_2021}, feedback covariate shift \citep{fannjiang_conformal_2022}, and survival analysis \citep{candes_conformalized_2023}, could be applied to WSCPS. Overall, our contributions offer a promising avenue for addressing covariate shifts in predictive modeling, with potential applications in diverse fields such as healthcare, finance, and climate science.

\acks{Part of this research was supported through the Flemish Government (AI Research Program).}

\bibliography{references}

\begin{thebibliography}{27}
\providecommand{\natexlab}[1]{#1}
\providecommand{\url}[1]{\texttt{#1}}
\expandafter\ifx\csname urlstyle\endcsname\relax
  \providecommand{\doi}[1]{doi: #1}\else
  \providecommand{\doi}{doi: \begingroup \urlstyle{rm}\Url}\fi

\bibitem[Boström(2022)]{bostrom_crepes_2022}
Henrik Boström.
\newblock crepes: a {Python} {Package} for {Generating} {Conformal} {Regressors} and {Predictive} {Systems}.
\newblock In \emph{Proceedings of the {Eleventh} {Symposium} on {Conformal} and {Probabilistic} {Prediction} with {Applications}}, pages 24--41. PMLR, August 2022.
\newblock URL \url{https://proceedings.mlr.press/v179/bostrom22a.html}.
\newblock ISSN: 2640-3498.

\bibitem[Boström et~al.(2021)Boström, Johansson, and Löfström]{bostrom_mondrian_2021}
Henrik Boström, Ulf Johansson, and Tuwe Löfström.
\newblock Mondrian conformal predictive distributions.
\newblock In \emph{Proceedings of the {Tenth} {Symposium} on {Conformal} and {Probabilistic} {Prediction} and {Applications}}, pages 24--38. PMLR, September 2021.
\newblock URL \url{https://proceedings.mlr.press/v152/bostrom21a.html}.

\bibitem[Candès et~al.(2023)Candès, Lei, and Ren]{candes_conformalized_2023}
Emmanuel~J. Candès, Lihua Lei, and Zhimei Ren.
\newblock Conformalized {Survival} {Analysis}, April 2023.
\newblock URL \url{http://arxiv.org/abs/2103.09763}.
\newblock arXiv:2103.09763 [stat].

\bibitem[Dua and Casey(2017)]{dua_uci_2017}
Dheeru Dua and Graff Casey.
\newblock {UCI} machine learning repository.
\newblock 2017.

\bibitem[Fannjiang et~al.(2022)Fannjiang, Bates, Angelopoulos, Listgarten, and Jordan]{fannjiang_conformal_2022}
Clara Fannjiang, Stephen Bates, Anastasios~N. Angelopoulos, Jennifer Listgarten, and Michael~I. Jordan.
\newblock Conformal prediction under feedback covariate shift for biomolecular design.
\newblock \emph{Proceedings of the National Academy of Sciences}, 119\penalty0 (43):\penalty0 e2204569119, October 2022.
\newblock \doi{10.1073/pnas.2204569119}.
\newblock URL \url{https://www.pnas.org/doi/abs/10.1073/pnas.2204569119}.
\newblock Publisher: Proceedings of the National Academy of Sciences.

\bibitem[Gibbs and Candes(2021)]{gibbs_adaptive_2021}
Isaac Gibbs and Emmanuel Candes.
\newblock Adaptive {Conformal} {Inference} {Under} {Distribution} {Shift}.
\newblock In \emph{Advances in {Neural} {Information} {Processing} {Systems}}, volume~34, pages 1660--1672. Curran Associates, Inc., 2021.
\newblock URL \url{https://proceedings.neurips.cc/paper/2021/hash/0d441de75945e5acbc865406fc9a2559-Abstract.html}.

\bibitem[Gibbs et~al.(2023)Gibbs, Cherian, and Candès]{gibbs_conformal_2023}
Isaac Gibbs, John~J. Cherian, and Emmanuel~J. Candès.
\newblock Conformal {Prediction} {With} {Conditional} {Guarantees}, December 2023.
\newblock URL \url{http://arxiv.org/abs/2305.12616}.
\newblock arXiv:2305.12616 [stat].

\bibitem[Gneiting and Raftery(2007)]{gneiting_strictly_2007}
Tilmann Gneiting and Adrian~E Raftery.
\newblock Strictly {Proper} {Scoring} {Rules}, {Prediction}, and {Estimation}.
\newblock \emph{Journal of the American Statistical Association}, 102\penalty0 (477):\penalty0 359--378, March 2007.
\newblock ISSN 0162-1459.
\newblock \doi{10.1198/016214506000001437}.
\newblock URL \url{https://doi.org/10.1198/016214506000001437}.
\newblock Publisher: Taylor \& Francis \_eprint: https://doi.org/10.1198/016214506000001437.

\bibitem[Gneiting et~al.(2007)Gneiting, Balabdaoui, and Raftery]{gneiting_probabilistic_2007}
Tilmann Gneiting, Fadoua Balabdaoui, and Adrian~E. Raftery.
\newblock Probabilistic forecasts, calibration and sharpness.
\newblock \emph{Journal of the Royal Statistical Society: Series B (Statistical Methodology)}, 69\penalty0 (2):\penalty0 243--268, 2007.
\newblock ISSN 1467-9868.
\newblock \doi{10.1111/j.1467-9868.2007.00587.x}.
\newblock URL \url{https://onlinelibrary.wiley.com/doi/abs/10.1111/j.1467-9868.2007.00587.x}.

\bibitem[Gretton et~al.(2008)Gretton, Smola, Huang, Schmittfull, Borgwardt, and Schölkopf]{gretton_covariate_2008}
Arthur Gretton, Alex Smola, Jiayuan Huang, Marcel Schmittfull, Karsten Borgwardt, and Bernhard Schölkopf.
\newblock Covariate {Shift} by {Kernel} {Mean} {Matching}.
\newblock In \emph{Dataset {Shift} in {Machine} {Learning}}. MIT Press, Cambridge, Mass., December 2008.
\newblock ISBN 978-0-262-25510-3.
\newblock URL \url{https://direct.mit.edu/books/edited-volume/3841/chapter/125883/Covariate-Shift-by-Kernel-Mean-Matching}.

\bibitem[Johansson et~al.(2023)Johansson, Löfström, and Boström]{johansson_conformal_2023}
Ulf Johansson, Tuwe Löfström, and Henrik Boström.
\newblock Conformal {Predictive} {Distribution} {Trees}.
\newblock \emph{Annals of Mathematics and Artificial Intelligence}, June 2023.
\newblock ISSN 1573-7470.
\newblock \doi{10.1007/s10472-023-09847-0}.
\newblock URL \url{https://doi.org/10.1007/s10472-023-09847-0}.

\bibitem[Jonkers et~al.(2024{\natexlab{a}})Jonkers, Avendano, Van~Wallendael, and Van~Hoecke]{jonkers_novel_2024}
Jef Jonkers, Diego~Nieves Avendano, Glenn Van~Wallendael, and Sofie Van~Hoecke.
\newblock A novel day-ahead regional and probabilistic wind power forecasting framework using deep {CNNs} and conformalized regression forests.
\newblock \emph{Applied Energy}, 361:\penalty0 122900, May 2024{\natexlab{a}}.
\newblock ISSN 0306-2619.
\newblock \doi{10.1016/j.apenergy.2024.122900}.
\newblock URL \url{https://www.sciencedirect.com/science/article/pii/S0306261924002836}.

\bibitem[Jonkers et~al.(2024{\natexlab{b}})Jonkers, Verhaeghe, Van~Wallendael, Duchateau, and Van~Hoecke]{jonkers_conformal_2024}
Jef Jonkers, Jarne Verhaeghe, Glenn Van~Wallendael, Luc Duchateau, and Sofie Van~Hoecke.
\newblock Conformal {Convolution} and {Monte} {Carlo} {Meta}-learners for {Predictive} {Inference} of {Individual} {Treatment} {Effects}, June 2024{\natexlab{b}}.
\newblock URL \url{http://arxiv.org/abs/2402.04906}.
\newblock arXiv:2402.04906 [cs, stat].

\bibitem[Kang and Schafer(2007)]{kang_demystifying_2007}
Joseph D.~Y. Kang and Joseph~L. Schafer.
\newblock Demystifying {Double} {Robustness}: {A} {Comparison} of {Alternative} {Strategies} for {Estimating} a {Population} {Mean} from {Incomplete} {Data}.
\newblock \emph{Statistical Science}, 22\penalty0 (4):\penalty0 523--539, November 2007.
\newblock ISSN 0883-4237, 2168-8745.
\newblock \doi{10.1214/07-STS227}.
\newblock URL \url{https://projecteuclid.org/journals/statistical-science/volume-22/issue-4/Demystifying-Double-Robustness--A-Comparison-of-Alternative-Strategies-for/10.1214/07-STS227.full}.

\bibitem[Papadopoulos et~al.(2002)Papadopoulos, Proedrou, Vovk, and Gammerman]{papadopoulos_inductive_2002}
Harris Papadopoulos, Kostas Proedrou, Volodya Vovk, and Alex Gammerman.
\newblock Inductive {Confidence} {Machines} for {Regression}.
\newblock In Tapio Elomaa, Heikki Mannila, and Hannu Toivonen, editors, \emph{Machine {Learning}: {ECML} 2002}, Lecture {Notes} in {Computer} {Science}, pages 345--356, Berlin, Heidelberg, 2002. Springer.
\newblock ISBN 978-3-540-36755-0.
\newblock \doi{10.1007/3-540-36755-1_29}.

\bibitem[Podkopaev and Ramdas(2021)]{podkopaev_distribution-free_2021}
Aleksandr Podkopaev and Aaditya Ramdas.
\newblock Distribution-free uncertainty quantification for classification under label shift.
\newblock In \emph{Proceedings of the {Thirty}-{Seventh} {Conference} on {Uncertainty} in {Artificial} {Intelligence}}, pages 844--853. PMLR, December 2021.
\newblock URL \url{https://proceedings.mlr.press/v161/podkopaev21a.html}.
\newblock ISSN: 2640-3498.

\bibitem[Prinster et~al.(2022)Prinster, Liu, and Saria]{prinster_jaws_2022}
Drew Prinster, Anqi Liu, and Suchi Saria.
\newblock {JAWS}: {Auditing} {Predictive} {Uncertainty} {Under} {Covariate} {Shift}.
\newblock \emph{Advances in Neural Information Processing Systems}, 35:\penalty0 35907--35920, December 2022.
\newblock URL \url{https://proceedings.neurips.cc/paper_files/paper/2022/hash/e944bacecce6b06374ac39b260348db0-Abstract-Conference.html}.

\bibitem[Reddi et~al.(2015)Reddi, Poczos, and Smola]{reddi_doubly_2015}
Sashank Reddi, Barnabas Poczos, and Alex Smola.
\newblock Doubly {Robust} {Covariate} {Shift} {Correction}.
\newblock \emph{Proceedings of the AAAI Conference on Artificial Intelligence}, 29\penalty0 (1), February 2015.
\newblock ISSN 2374-3468.
\newblock \doi{10.1609/aaai.v29i1.9576}.
\newblock URL \url{https://ojs.aaai.org/index.php/AAAI/article/view/9576}.

\bibitem[Shafer and Vovk(2008)]{shafer_tutorial_2008}
Glenn Shafer and Vladimir Vovk.
\newblock A {Tutorial} on {Conformal} {Prediction}.
\newblock \emph{Journal of Machine Learning Research}, 9\penalty0 (12):\penalty0 371--421, 2008.
\newblock ISSN 1533-7928.
\newblock URL \url{http://jmlr.org/papers/v9/shafer08a.html}.

\bibitem[Tibshirani et~al.(2019)Tibshirani, Foygel~Barber, Candes, and Ramdas]{tibshirani_conformal_2019}
Ryan~J Tibshirani, Rina Foygel~Barber, Emmanuel Candes, and Aaditya Ramdas.
\newblock Conformal {Prediction} {Under} {Covariate} {Shift}.
\newblock In \emph{Advances in {Neural} {Information} {Processing} {Systems}}, volume~32. Curran Associates, Inc., 2019.
\newblock URL \url{https://proceedings.neurips.cc/paper/2019/hash/8fb21ee7a2207526da55a679f0332de2-Abstract.html}.

\bibitem[Vovk(2022)]{vovk_universal_2022}
Vladimir Vovk.
\newblock Universal predictive systems.
\newblock \emph{Pattern Recognition}, 126:\penalty0 108536, June 2022.
\newblock ISSN 0031-3203.
\newblock \doi{10.1016/j.patcog.2022.108536}.
\newblock URL \url{https://www.sciencedirect.com/science/article/pii/S0031320322000176}.

\bibitem[Vovk et~al.(2018)Vovk, Nouretdinov, Manokhin, and Gammerman]{vovk_conformal_2018}
Vladimir Vovk, Ilia Nouretdinov, Valery Manokhin, and Alex Gammerman.
\newblock Conformal {Predictive} {Distributions} with {Kernels}.
\newblock In Lev Rozonoer, Boris Mirkin, and Ilya Muchnik, editors, \emph{Braverman {Readings} in {Machine} {Learning}. {Key} {Ideas} from {Inception} to {Current} {State}: {International} {Conference} {Commemorating} the 40th {Anniversary} of {Emmanuil} {Braverman}'s {Decease}, {Boston}, {MA}, {USA}, {April} 28-30, 2017, {Invited} {Talks}}, Lecture {Notes} in {Computer} {Science}, pages 103--121. Springer International Publishing, Cham, 2018.
\newblock ISBN 978-3-319-99492-5.
\newblock \doi{10.1007/978-3-319-99492-5_4}.
\newblock URL \url{https://doi.org/10.1007/978-3-319-99492-5_4}.

\bibitem[Vovk et~al.(2019)Vovk, Shen, Manokhin, and Xie]{vovk_nonparametric_2019}
Vladimir Vovk, Jieli Shen, Valery Manokhin, and Min-Ge Xie.
\newblock Nonparametric predictive distributions based on conformal prediction.
\newblock \emph{Machine Language}, 108\penalty0 (3):\penalty0 445--474, March 2019.
\newblock ISSN 0885-6125.
\newblock \doi{10.1007/s10994-018-5755-8}.
\newblock URL \url{https://doi.org/10.1007/s10994-018-5755-8}.

\bibitem[Vovk et~al.(2020{\natexlab{a}})Vovk, Petej, Nouretdinov, Manokhin, and Gammerman]{vovk_computationally_2020}
Vladimir Vovk, Ivan Petej, Ilia Nouretdinov, Valery Manokhin, and Alexander Gammerman.
\newblock Computationally efficient versions of conformal predictive distributions.
\newblock \emph{Neurocomputing}, 397:\penalty0 292--308, July 2020{\natexlab{a}}.
\newblock ISSN 0925-2312.
\newblock \doi{10.1016/j.neucom.2019.10.110}.
\newblock URL \url{https://www.sciencedirect.com/science/article/pii/S0925231219316042}.

\bibitem[Vovk et~al.(2020{\natexlab{b}})Vovk, Petej, Toccaceli, Gammerman, Ahlberg, and Carlsson]{vovk_conformal_2020}
Vladimir Vovk, Ivan Petej, Paolo Toccaceli, Alexander Gammerman, Ernst Ahlberg, and Lars Carlsson.
\newblock Conformal calibrators.
\newblock In \emph{Proceedings of the {Ninth} {Symposium} on {Conformal} and {Probabilistic} {Prediction} and {Applications}}, pages 84--99. PMLR, August 2020{\natexlab{b}}.
\newblock URL \url{https://proceedings.mlr.press/v128/vovk20a.html}.
\newblock ISSN: 2640-3498.

\bibitem[Vovk et~al.(2022)Vovk, Gammerman, and Shafer]{vovk_algorithmic_2022}
Vladimir Vovk, Alexander Gammerman, and Glenn Shafer.
\newblock \emph{Algorithmic {Learning} in a {Random} {World}}.
\newblock Springer International Publishing, Cham, 2022.
\newblock ISBN 978-3-031-06648-1 978-3-031-06649-8.
\newblock \doi{10.1007/978-3-031-06649-8}.
\newblock URL \url{https://link.springer.com/10.1007/978-3-031-06649-8}.

\bibitem[Yang et~al.(2022)Yang, Kuchibhotla, and Tchetgen]{yang_doubly_2022}
Yachong Yang, Arun~Kumar Kuchibhotla, and Eric~Tchetgen Tchetgen.
\newblock Doubly {Robust} {Calibration} of {Prediction} {Sets} under {Covariate} {Shift}, December 2022.
\newblock URL \url{http://arxiv.org/abs/2203.01761}.
\newblock arXiv:2203.01761 [math, stat].

\end{thebibliography}

\appendix

\section{Split Conformal Predictive System}\label{apd:scps}
For split CPS (SCPS), the same procedure is followed as a split conformal prediction; the training sequence $z_{1:n}$ is split into two: a proper training sequence $z_{1:m}$ and calibration sequence $z_{m+1:n}$. Similarly as an CPS, an SCPS is defined as a function that is both a split conformal transducer (Definition \ref{def:sct}) and an RPS (Definition \ref{def:RPS}) \citep{vovk_computationally_2020}.
\begin{definition}[Inductive (Split) Conformity Measure, \cite{vovk_algorithmic_2022}]
\label{def:split-measure}
A split conformity measure is a measurable function $A: Z^m \times Z \rightarrow \mathbb{R}$ that is invariant with respect to permutations of the proper training sequence $z_{1:m}$.
\end{definition}
\begin{definition}[Split Conformal Transducer, \cite{vovk_computationally_2020}]
    \label{def:sct}
    The split conformal transducer determined by a split conformity measure $A$ (see Definition \ref{def:split-measure}) is defined as,
    \begin{equation}
        \begin{split}
            Q(z_1,...,z_n, (x_{n+1},y), \tau) :=& \sum_{i=m+1}^{n} [R_i < R^y_{n+1}] \frac{1}{n-m+1} \\
            &+ \sum_{i=m+1}^{n} [R_i = R^y_{n+1}] \frac{\tau}{n-m+1} \\
            &+ \frac{\tau}{n-m+1}
        \end{split}
    \end{equation}
    where conformity scores $R_i$ and $R^y$ are defined by
    \begin{equation*}
        \begin{split}
            R_i &:= A(z_1,...,z_m,(x_i,y_i)), \qquad i=m+1,...,n, \\
            R^y_{n+1} &:= A(z_1,...,z_m,(x_{n+1},y_{n+1})), \qquad y \in \mathbb{R}.
        \end{split}
    \end{equation*}
\end{definition}

\cite{vovk_computationally_2020} proofs that any split conformal transducer is an RPS if and only if it is based on a balanced isotonic split conformity measure (Definition \ref{def:bal-iso-measure}).
\begin{definition}[Isotonic Split Conformity Measure, \cite{vovk_computationally_2020}]
\label{def:iso-measure}
    A split conformity measure $A$ is isotonic if, for all $m$, $z_{1:m}$, and $x$, $A(z_1, ...,z_m, (x,y))$ is isotonic in $y$, i.e.,
    \[
    y \leq y' \Rightarrow A(z_1,...,z_m,(x,y)) \leq A(z_1, ...,z_m,(x,y'))
    \]
\end{definition}
\begin{definition}[Balanced Isotonic Split Conformity Measure, \cite{vovk_computationally_2020}]
\label{def:bal-iso-measure}
    An isotonic split conformity measure $A$ (see Definition \ref{def:iso-measure}) is balanced if, for any $m$ and $z_1, ..., z_m$, the set 
    \begin{equation*}
        \text{conv } A(z_1,...,z_m, (x, \mathbb{R})) := \text{conv } \{A(z_1,...,z_m, (x, y))| y \in \mathbb{R}\}
    \end{equation*}
    where conv stands for the convex closure in $\mathbb{R}$.
\end{definition}

\section{Python Package: \texttt{crepes-weighted}}
\label{apd:package}
For the simulation experiments in this work, we implemented the proposed WSCPS and the WCP \citep{tibshirani_conformal_2019} in \texttt{crepes-weighted}, which is an extension of \texttt{crepes} \citep{bostrom_crepes_2022}, a Python package that implements conformal classifiers, regressors, and predictive systems on top of any standard classifier and regressor. \texttt{crepes-weighted} relies on the same classes and functions as \texttt{crepes}, with the slight modification that for the \texttt{ConformalRegressor} and \texttt{ConformalPredictiveSystem} classes, the methods \texttt{fit} and \texttt{predict} needs to include the likelihood ratios of each calibration and test object respectively.

The source code of \texttt{crepes-weighted} is made open-source and can be found at \url{https://github.com/predict-idlab/crepes-weighted}.
\end{document}